%% file: main.tex
\documentclass[11pt]{article}
\PassOptionsToPackage{table}{xcolor}
\usepackage[preprint]{acl}
\usepackage{amssymb}
\usepackage{bm}
\usepackage{times}
\usepackage{latexsym}
\usepackage[T1]{fontenc}
\usepackage[utf8]{inputenc}
\usepackage{microtype}
\usepackage{inconsolata}
\usepackage{graphicx}
\usepackage{amsmath}
\usepackage{multirow}
\usepackage{tikz}
\usepackage{xcolor}
\usepackage{booktabs} 
\usepackage{amsmath}
\definecolor{baselinerow}{gray}{0.92}   
\title{UAPAR: Uncertainty-Aware Pedestrian Attribute Recognition via Evidential Deep Learning}
\definecolor{mypurple}{RGB}{112, 48, 160}
\author{Zhuofan Lou\thanks{\ \ Equal contribution.} \quad 
  Shihang Zhang\footnotemark[1] \quad 
  Fangle Zhu
  \quad
  Shengjie Ye
  \quad
  Pingyu Wang\thanks{\ \ Corresponding author.} \\
  College of Electronics and Information Engineering, Sichuan University, China \\
  \texttt{\{louzhuofan, shihang\_zhang\}@stu.scu.edu.cn} \\
  \texttt{wangpingyu@scu.edu.cn} \\}

\begin{document}

\maketitle

\input{latex/abstract}

\section{Introduction}
\input{latex/introduction}
\section{Related Work}
\input{latex/related_work}
\section{Method} 
\input{latex/method}
\section{Experiments}

\input{latex/experiment}

\section{Conclusion} 
\input{latex/conclusion}
\section{Acknowledgements}
\input{latex/acknowledge}

 \bibliography{references}

\end{document}

%% file: latex/abstract.tex
\begin{abstract}
We propose \textbf{UAPAR}, an \textbf{U}ncertainty-\textbf{A}ware \textbf{P}edestrian \textbf{A}ttribute \textbf{R}ecognition framework. To the best of our knowledge, this is the first EDL-based uncertainty-aware framework for pedestrian attribute recognition (PAR). Unlike conventional deterministic methods, which fail to assess prediction reliability on low-quality samples, UAPAR effectively identifies unreliable predictions and thus enhances system robustness in complex real-world scenarios. To achieve this, UAPAR incorporates Evidential Deep Learning (EDL) into a CLIP-based architecture. Specifically, a Region-Aware Evidence Reasoning module employs cross-attention and spatial prior masks to capture fine-grained local features, which are further processed by an evidence head to estimate attribute-wise epistemic uncertainty. To further enhance training robustness, we develop an uncertainty-guided dual-stage curriculum learning strategy to alleviate the adverse effects of severe label noise during training. Extensive experiments on the PA100K, PETA, RAPv1, and RAPv2 datasets demonstrate that UAPAR achieves competitive or superior performance. Furthermore, qualitative results confirm that the proposed framework generates uncertainty estimates that are predictive of challenging or erroneous samples.
\end{abstract}

%% file: latex/introduction.tex
\begin{figure}[t]
    \centering
    \includegraphics[width=\linewidth]{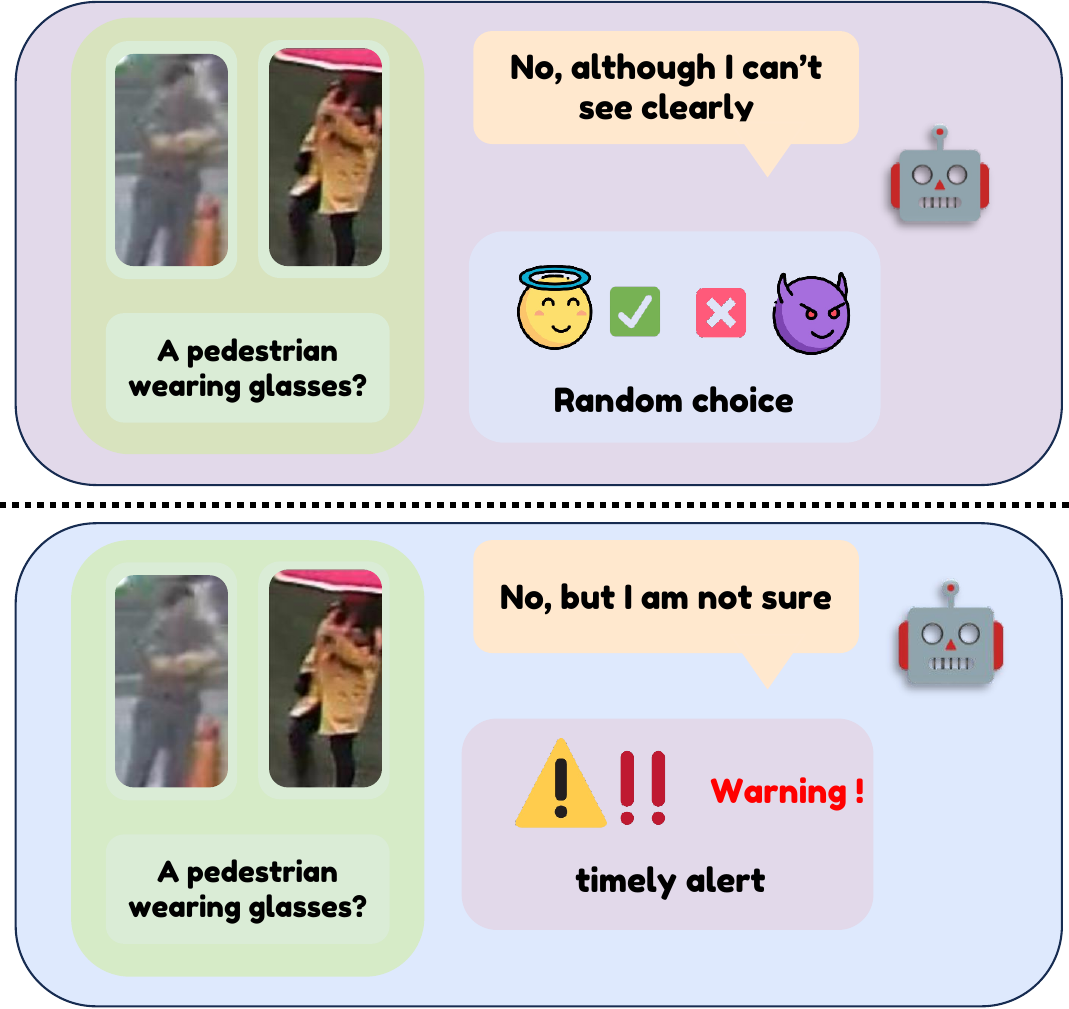}
    \caption{Illustration of prediction paradigms under severe occlusion. Lacking visual evidence, the traditional model (top) makes an unfounded random choice, whereas the ideal model (bottom) accurately quantifies epistemic uncertainty to provide a timely alert.}
    \label{fig:model_compare}
\end{figure}

Pedestrian Attribute Recognition (PAR) aims to identify pedestrian semantic attributes from surveillance images. Although many outstanding works utilize deep learning and pre-training methods ~\cite{9782406,NEURIPS2023_9ed1c94a, 10664464}, they generally adopt the point estimation paradigm, i.e., outputting a single probability value via sigmoid or softmax. This paradigm fails to distinguish credible predictions supported by evidence from random guesses lacking visual cues. As illustrated in Figure~\ref{fig:model_compare}, when facing challenging images with severe occlusion or incomplete capture, the upper traditional model may be forced to make unfounded random choices. In contrast, the lower ideal model can accurately quantify epistemic uncertainty alongside its predictions, providing a timely alert when visual evidence is lacking. Therefore, endowing PAR models with the ability to evaluate prediction reliability is particularly urgent.

Recent PAR studies ~\cite{shen2023sspnetscalespatialpriors,9782406,YiboZhou2025AST} have mainly focused on improving feature extraction and context modeling capabilities, achieving significant performance improvements on major benchmark datasets. However, existing methods generally lack the ability to measure prediction uncertainty, and this lack of reliability awareness directly leads the model to fit all data indiscriminately. When encountering extreme samples without visual cues due to severe occlusion, forced fitting not only causes the model to overfit on noise but also interferes with the learning of effective representations. Therefore, existing works urgently need a mechanism to perceive sample quality and adjust the learning pace.

To address these issues, we propose the UAPAR framework. At the architecture level, we design a Region-Aware Evidence Reasoning mechanism, forcing the model to focus on local body regions corresponding to attributes. Based on the extracted evidential features, we introduce Evidential Deep Learning and model predictions as Beta distributions, enabling the model to simultaneously output expected prediction probabilities and epistemic uncertainty. At the training level, we propose an uncertainty-based two-stage progressive training paradigm. Combined with a Curriculum Learning strategy (CL), it guides the model to transition from simple samples to hard samples based on its own cognitive boundaries, avoiding forced fitting on meaningless occluded samples.

Our main contributions are as follows:
\begin{itemize}
    \item We propose the first uncertainty-aware framework for pedestrian attribute recognition. By introducing Evidential Deep Learning and Beta distributions, our model outputs expected probabilities and uncertainty.
    \item We design a Region-Aware Evidence Reasoning mechanism guiding the model to focus on reliable local visual evidence, providing a feature foundation for uncertainty evaluation.
    \item We propose an uncertainty-based two-stage progressive training paradigm integrated with a curriculum learning strategy (CL). This achieves robust easy-to-hard learning based on the cognitive boundaries of the model, preventing fitting on meaningless samples.
    \item Extensive experiments demonstrate superior performance on mainstream benchmarks and verify our model generates meaningful uncertainty metrics for difficult samples.
\end{itemize}

%% file: latex/related_work.tex
\subsection{Pedestrian Attribute Recognition}

Pedestrian Attribute Recognition (PAR) methodology has evolved from CNN-based multi-task learning~\cite{Zhang_2014_CVPR, 7254184, Wang_2017_ICCV} and Transformer-based global dependency modeling~\cite{Zhao_Sang_Ding_Han_Di_Yan_2019, TANG2022159, 10148996} to recent vision-language pre-training paradigms utilizing prompt fusion~\cite{9782406, NEURIPS2023_9ed1c94a, 10664464}. Despite steady accuracy improvements, these generations remain trapped in a deterministic classification paradigm. Each attribute yields a single probability, rendering models incapable of quantifying epistemic uncertainty when facing occluded or ambiguous samples, thus causing overconfident predictions. This deterministic bottleneck motivates our introduction of evidential reasoning to the PAR pipeline.

\begin{figure*}[t]
    \centering
    \includegraphics[width=\linewidth]{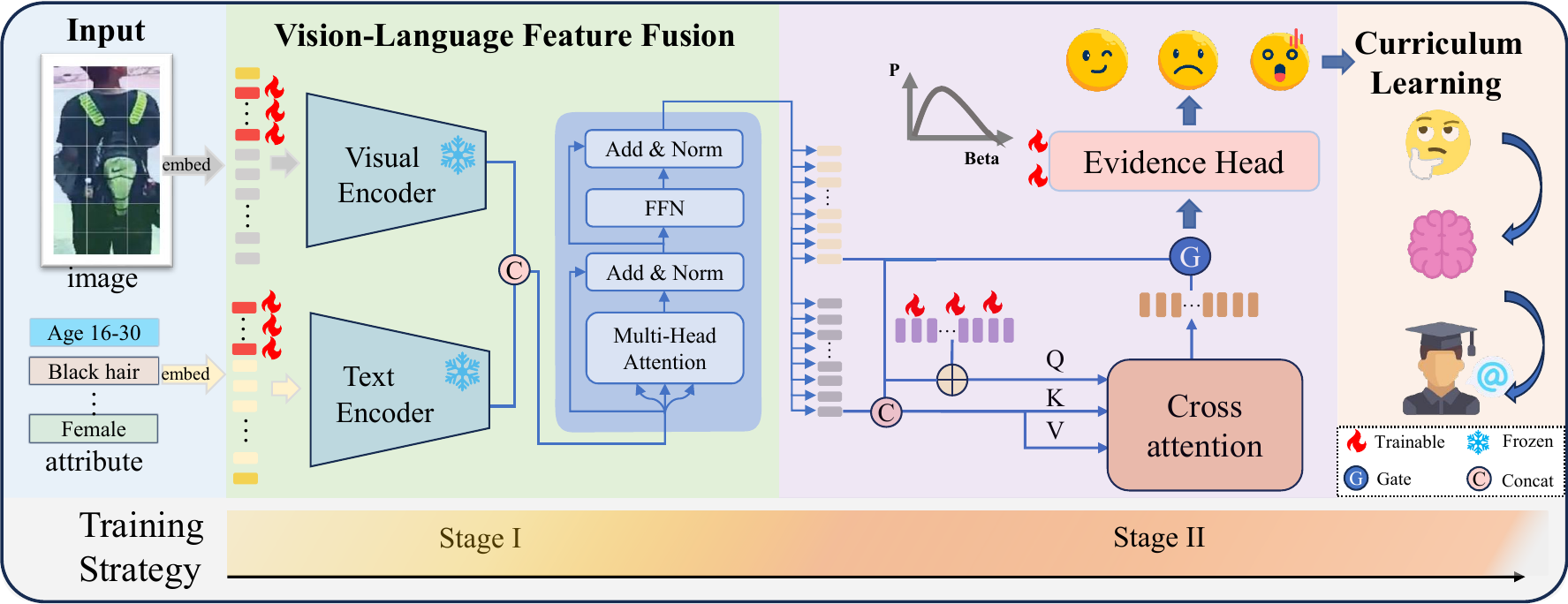}
    \caption{The pipeline of the proposed method. Cross-modal features are first extracted and fused via prompt-tuned dual encoders and a Multi-Modal Fusion module. A cross-attention mechanism then performs region-aware localization, and an evidence head models predictions as Beta distributions to quantify uncertainty. Finally, this uncertainty guides progressive curriculum learning across a two-stage training strategy.}
    \label{fig:pipeline}
\end{figure*}

\subsection{Evidential Deep Learning}

Breaking this bottleneck requires explicitly quantifying missing evidence. Evidential Deep Learning (EDL)~\cite{NEURIPS2018_a981f2b7} addresses this by placing a Dirichlet prior over categorical distributions to produce per-class evidence and a vacuity score. For multi-label PAR, this reduces to independent Beta distributions. While EDL is widely adopted in medical imaging, action recognition, and autonomous driving~\cite{zou2025reliablemedicalimagesegmentation, he2024eplevidentialprototypelearning, Bao_2021_ICCV, Chen_2023_CVPR, 10341826, durasov2024uncertaintyestimation3dobject, 11217233}, it has never been applied to PAR. Our framework fills this gap by combining Beta evidential heads with region-aware cross-attention, grounding per-attribute uncertainty quantification in spatial evidence to handle ambiguous and low-resolution pedestrians.

\subsection{Curriculum Learning}

Feeding this uncertainty signal back into training naturally aligns with Curriculum Learning (CL)~\cite{10.1145/1553374.1553380}. Traditional CL and Self-Paced Learning~\cite{10.5555/2997189.2997322, 10.5555/2886521.2886696, pmlr-v80-jiang18c} rely on surrogate difficulty metrics that conflate hard-but-learnable samples with inherently noisy ones. The vacuity score from our Beta evidential head provides a principled alternative where low uncertainty indicates learned samples, moderate uncertainty marks the learning frontier, and high uncertainty signals genuine ambiguity. Accordingly, we design a two-phase curriculum that utilizes binary cross-entropy loss pacing during warmup and transitions to Gaussian uncertainty weighting, replacing heuristic schedules with adaptive, evidence-driven optimization.

%% file: latex/method.tex
\subsection{Overall Framework}

As illustrated in Figure~\ref{fig:pipeline}, our framework comprises three core modules: (1) Vision-Language Feature Fusion adapts CLIP encoders via learnable prompts and cross-modal interaction to extract attribute-aware unified representations. (2) Evidential Uncertainty Estimation models attribute predictions as Beta distributions via an evidence head for principled uncertainty quantification. (3) Uncertainty-based Curriculum Learning utilizes this uncertainty to guide training, enabling progressive easy-to-hard learning. Finally, a two-stage training strategy warms up with BCE loss to establish stable features (Stage I), then switches to Beta evidential loss with gradual regularization (Stage II).

\subsection{Vision-Language Feature Fusion}
Given a pedestrian image $I \in \mathbb{R}^{H \times W \times 3}$, where $H$ and $W$ denote the image height and width respectively, and $N$ denotes textual attribute descriptions, we extract features using pre-trained CLIP~\cite{radford2021learningtransferablevisualmodels} vision and text encoders.

\textbf{Vision Branch:} The image is divided into non-overlapping patches and processed by the Transformer vision encoder, yielding patch-level representations $\mathbf{V} = \{\mathbf{v}_{\text{cls}}, \mathbf{v}_1, \ldots, \mathbf{v}_P\} \in \mathbb{R}^{(P+1) \times d_v}$, where $P$ denotes the patch count and $d_v$ is the visual feature dimension.

\textbf{Text Branch:} For each attribute $j$, the phrase is expanded into a descriptive sentence template (e.g., "Hat" $\rightarrow$ "A pedestrian wearing a hat") and processed by the CLIP text encoder to extract attribute-level text features $\mathbf{T} = \{\mathbf{t}_1, \ldots, \mathbf{t}_N\} \in \mathbb{R}^{N \times d_t}$, where $d_t$ represents the text feature dimension.

To adapt the pre-trained CLIP encoders for pedestrian attribute recognition, inspired by PromptPAR~\cite{wang2024pedestrianattributerecognitionclip}, we introduce Deep Prompt Tuning into both encoders. In the first $D_v$ vision Transformer layers, learnable prompt vectors $\{\mathbf{p}_1^{(l)}, \ldots, \mathbf{p}_M^{(l)}\}$ are inserted between the class (\textit{cls}) token and patch features, where $M$ is the number of prompt vectors and $l$ denotes the layer index. Similarly, learnable text prompts are inserted into every text Transformer layer. These vectors interact with original tokens via layer-wise self-attention, injecting task-specific knowledge without altering pre-trained CLIP weights, thus guiding the encoders to focus on relevant feature patterns.

To capture visual-textual interactions, we concatenate both features along the token dimension and input them into the Multi-Modal Fusion module:
\begin{equation}
    \mathbf{F} = \text{MultiModalFusion}([\mathbf{T}'; \mathbf{V}'])
\end{equation}
where $\mathbf{T}' = \text{Proj}_t(\mathbf{T}) \in \mathbb{R}^{N \times d}$ and $\mathbf{V}' = \text{Proj}_v(\mathbf{V}) \in \mathbb{R}^{(P+1) \times d}$ are features projected to a common dimension $d$ via linear projection layers $\text{Proj}_t$ and $\text{Proj}_v$, respectively, and $[\cdot;\cdot]$ denotes concatenation. This module consists of stacked Transformer blocks containing multi-head self-attention, layer normalization, and feed-forward networks. Self-attention enables both modalities to acquire rich local context, establishing preliminary associations between attributes and image regions. The complete output is denoted as $\mathbf{F} \in \mathbb{R}^{L \times d}$ with $L$ total tokens, where the first $N$ text tokens correspond to the attribute-aware fused features $\mathbf{F}_{\text{attr}} \in \mathbb{R}^{N \times d}$.

\subsection{Region-Aware Evidence Reasoning}
Although Multi-Modal Fusion establishes preliminary local associations, self-attention treats all spatial regions equally. Due to the prevalent long-tail distribution in real-world datasets, the model easily learns spurious correlations lacking causality. To address this, we introduce a cross-attention module. By incorporating a learnable attribute query matrix, this module dynamically updates and explicitly guides attributes to precisely locate causal image regions.

Specifically, a learnable attribute query matrix $\mathbf{W}_{\text{attr}} \in \mathbb{R}^{N \times d}$ is added to the fused features, serving as an independent query:
\begin{equation}
    \mathbf{Q} = \text{LN}(\mathbf{W}_{\text{attr}} + \mathbf{F}_{\text{attr}})
\end{equation}
Key and Value are obtained by projecting the complete fusion output: $\mathbf{K} = \mathbf{F} \mathbf{W}_K,\ \mathbf{V} = \mathbf{F} \mathbf{W}_V$. Multi-head cross-attention scores are computed as:
\begin{equation}
    \mathbf{A} = \frac{\mathbf{Q} \mathbf{W}_Q \cdot \mathbf{K}^\top}{\sqrt{d_h}}
\end{equation}
where $d_h = d / H$ is the dimension per head, and $H$ denotes the number of heads.

\textbf{Spatial Prior Mask (SPM):} To assist precise localization and inject human structural priors, we design a spatial prior mask $\mathbf{M} \in \mathbb{R}^{N \times L}$. For each attribute $j$ and spatial location $l$, it is defined as:
\begin{equation}
    M_{j,l} = \begin{cases}
        +\gamma, & \text{if } l \text{ is in a relevant region}, \\
        -\gamma, & \text{if } l \text{ is irrelevant}, \\
        0, & \text{for global attributes}.
    \end{cases}
\end{equation}
where $\gamma$ is a learnable parameter. This mask is added to the attention scores before the softmax operation:
\begin{equation}
    \hat{\mathbf{A}} = \text{softmax}(\mathbf{A} + \mathbf{M})
\end{equation}
This soft bias helps the model learn useful associations while adjusting attention. Based on these prior-injected scores, the module extracts region-aware features $\mathbf{r}_j \in \mathbb{R}^d$ for each attribute via weighted aggregation:
\begin{equation}
    \mathbf{r}_j = \sum_{l=1}^{L} \hat{A}_{j,l} \cdot \mathbf{V}_l
\end{equation}

\textbf{Gated Fusion:} To enhance causal feature contributions and suppress non-causal interference, we employ a gating mechanism to adaptively fuse region-aware features $\mathbf{r}_j$ with the original fused features $\mathbf{F}_{\text{attr},j}$:
\begin{equation}
    g_j = \sigma(\mathbf{W}_g [\mathbf{F}_{\text{attr},j}; \mathbf{r}_j] + b_g)
\end{equation}
\begin{equation}
    \tilde{\mathbf{f}}_j = g_j \cdot \mathbf{r}_j + (1 - g_j) \cdot \mathbf{F}_{\text{attr},j}
\end{equation}
where $g_j$ represents the gating weight for the $j$-th attribute, $\mathbf{W}_g$ is a learnable weight matrix, $\sigma$ is the sigmoid function, and $[\cdot;\cdot]$ denotes concatenation. Initializing the gating bias $b_g$ to a negative value ensures the model relies on original features during early training, gradually incorporating regional evidence later.

\subsection{Evidential Uncertainty Estimation}
Traditional methods output direct probability point estimates via a sigmoid classifier, lacking uncertainty measurement. To address this, we adopt the Evidential Deep Learning (EDL) framework, predicting a distribution over probabilities rather than single point values. 

Since pedestrian attribute recognition is a binary classification task, the Dirichlet distribution naturally degrades to a Beta distribution. The true probability $\theta_j \in [0,1]$ for each attribute is unknown, and its class probability prior is modeled as:
\begin{equation}
    p(\theta_j \mid \alpha_j, \beta_j) = \text{Beta}(\theta_j; \alpha_j, \beta_j)
\end{equation}
where $\alpha_j$ and $\beta_j$ are the parameters of the Beta distribution. The fused features $\tilde{\mathbf{f}}_j$ are mapped into two non-negative evidence values via a linear layer and Softplus activation:
\begin{equation}
    (\varepsilon_j^+, \varepsilon_j^-) = \text{Softplus}(\mathbf{W}_e \tilde{\mathbf{f}}_j + \mathbf{b}_e)
\end{equation}
where $\varepsilon_j^+$ and $\varepsilon_j^-$ represent evidence supporting the presence and absence of the attribute respectively, $\mathbf{W}_e$ is the learnable weight matrix, and $\mathbf{b}_e$ is the bias vector of the linear layer. Following EDL theory, Beta distribution parameters are defined as $\alpha_j = \varepsilon_j^+ + 1$ and $\beta_j = \varepsilon_j^- + 1$. The predicted probability $\hat{p}_j$ (Beta expectation) and vacuity uncertainty $u_j$ are:
\begin{equation}
    \hat{p}_j = \frac{\alpha_j}{\alpha_j + \beta_j}, \quad u_j = \frac{2}{\alpha_j + \beta_j}
\end{equation}
where $\hat{p}_j$ estimates the presence probability of attribute $j$, and $u_j \in (0, 1]$ quantifies evidence scarcity. With minimal evidence, the Beta distribution approaches uniformity ($u_j \approx 1$); as evidence accumulates, it concentrates, shifting $u_j$ toward $0$.

\subsection{Two-Stage Curriculum Training}

\begin{figure}[t]
    \centering
    \includegraphics[width=\linewidth]{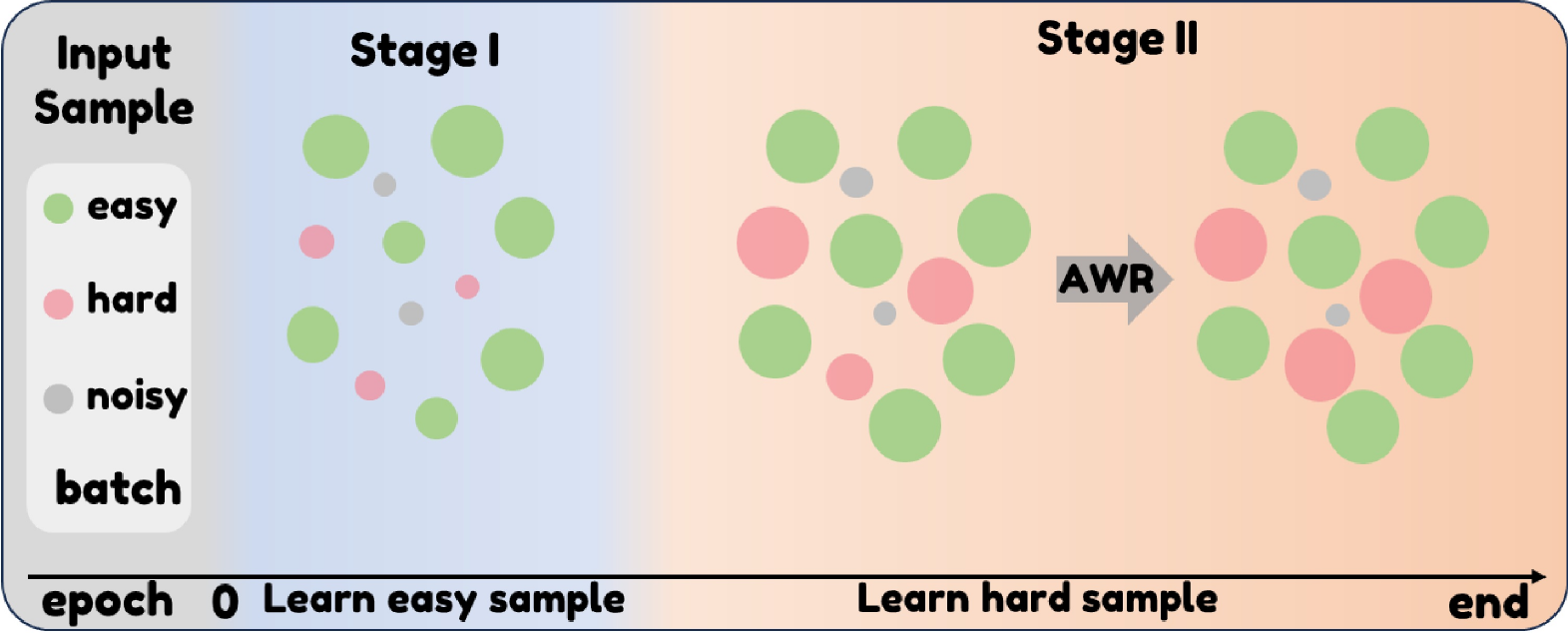}
    \caption{Two-stage curriculum training strategy. Stage I prioritizes CE loss-ranked easy samples, expanding the pacing boundary. Stage II targets EDL uncertainty-identified boundary samples, applying Attribute-Weighted Regularization (AWR) to refine evidence. Circle size reflects weight.}
    \label{fig:clu}
\end{figure}

We introduce a two-stage curriculum learning mechanism for a smooth transition from loss-guided warm-up to uncertainty-guided evidential learning, as shown in Figure~\ref{fig:clu}.

In Stage I, we train the model for $E_w$ epochs. We employ binary cross-entropy (BCE) loss to build initial features:
\begin{equation}
    \mathcal{L}_{\text{BCE}} = -\frac{1}{N}\sum_{j=1}^{N} \left[y_j \log \hat{p}_j + (1-y_j)\log(1-\hat{p}_j)\right]
\end{equation}
Here, BCE loss directly measures sample difficulty. Within each mini-batch, simpler samples with lower losses receive higher weights, guiding the model to prioritize reliable visual patterns.

For Stage II, training switches to the Beta evidential loss from epoch $E_w{+}1$. The model now outputs meaningful per-attribute vacuity uncertainty $u_j$, allowing us to use sample-level mean uncertainty $\bar{u}_i = \frac{1}{N}\sum_j u_{i,j}$, where $u_{i,j}$ is the vacuity uncertainty of the $j$-th attribute for the $i$-th sample, as the core difficulty indicator. The $\bar{u}_i$ distribution objectively characterizes the learning state of a sample: low values indicate well-fitted simple samples, moderate values represent the informative learning frontier, and high values typically denote unreliable noisy data. To focus on the learning frontier while avoiding overfitting to simple or noisy samples, we employ a Gaussian weighting function:
\begin{equation}
    w_i = \exp\!\left(-\frac{(\bar{u}_i - c(t))^2}{2\,\sigma^2(t)}\right)
\end{equation}
where $w_i$ denotes the weight assigned to the $i$-th sample, and $\sigma(t)$ is the standard deviation controlling the width of the Gaussian function. The center parameter $c(t)$ shifts dynamically from low to high during training to track the evolving frontier.

The evidential loss follows a standard hybrid paradigm:
\begin{equation}
    \mathcal{L}_{\text{EDL}} = \mathcal{L}_{\text{CE-BR}} + \lambda_t \cdot \mathcal{L}_{\text{MSE-BR}}
\end{equation}
The classification term $\mathcal{L}_{\text{CE-BR}}$ shares an identical mathematical formulation with the BCE in Stage I, but $\hat{p}_j$ now represents the Beta expectation, ensuring stable gradients. The calibration term $\mathcal{L}_{\text{MSE-BR}}$ penalizes prediction errors and variances to actively accumulate evidence, with total evidence strength $S_j = \alpha_j + \beta_j$:
\begin{equation}
    \mathcal{L}_{\text{MSE-BR}} = \frac{1}{N}\sum_{j=1}^{N}\left[(y_j - \hat{p}_j)^2 + \frac{\hat{p}_j(1-\hat{p}_j)}{S_j + 1}\right]
\end{equation}
where $y_j$ represents the ground-truth label for the $j$-th attribute. During training, weight $\lambda_t$ linearly increases from $0$ to $\lambda_{\max}$ to gradually sharpen the evidence distribution.

Regarding Attribute-Weighted Regularization (AWR), a standard evidential loss is easily dominated by common samples later in training, stalling optimization for few-shot attributes. We introduce a directional penalty term to enhance minority-class feature mining:
\begin{equation}
    \mathcal{L}_{\text{AWR}} = \frac{1}{N}\sum_{j=1}^{N}\left[y_j \cdot \varepsilon_j^{-} + (1 - y_j) \cdot \varepsilon_j^{+}\right]
\end{equation}
This term restricts incorrect prediction signals, preventing the network from bypassing hard samples via high uncertainty. It smooths late-stage gradient oscillations and forces the model to learn long-tail features, mitigating data imbalance biases. The final Stage II objective is:
\begin{equation}
    \mathcal{L}_{\text{Stage\,II}} = w_i \cdot \mathcal{L}_{\text{EDL}} + \lambda_{\text{awr}} \cdot \mathcal{L}_{\text{AWR}}
\end{equation}
where $\lambda_{\text{awr}}$ is the trade-off weight balancing the AWR loss term.

%% file: latex/experiment.tex
\input{table/compare_with_sota}

\subsection{Datasets and Implementation Details}
To validate the effectiveness of our approach, we conduct experiments on four mainstream pedestrian attribute recognition benchmarks: PETA~\cite{10.1145/2647868.2654966}, PA100K~\cite{liu2017hydraplusnetattentivedeepfeatures}, RAPv1~\cite{li2016richlyannotateddatasetpedestrian}, and RAPv2~\cite{8510891}. For brevity, detailed dataset statistics and further implementation specifics are deferred to the supplementary material.

The ViT-L/14 version of CLIP is adopted as the visual backbone, where original images are padded to $224 \times 224$ resolution. All models are trained on NVIDIA RTX 4090 GPUs using SGD with a base lr of $6 \times 10^{-3}$, a CLIP lr of $4 \times 10^{-3}$, weight decay $5 \times 10^{-4}$, and effective batch size 48. Training follows a two-stage schedule: Stage I from epoch 1 to $E_w$ with $E_w=12$ uses BCE loss with loss-based curriculum pacing; Stage II from epoch $E_{w+1}$ to end employs Beta Evidential Loss with uncertainty-based Gaussian curriculum weighting.

\subsection{Comparison with the SOTA Methods}
\input{table/zero_shot}
\paragraph{Superiority in Comprehensive Recognition} As summarized in Table \ref{tab:1},\ref{tab:2}, the proposed UAPAR establishes new state-of-the-art results in mean Accuracy across the PA100K (88.48\%) and PETA (90.74\%) datasets. This dominance in class-level recognition extends significantly to the RAP benchmarks, where our approach achieves the highest mean Accuracy of 87.48\% on RAPv1 and 85.71\% on RAPv2, surpassing recent competitive methods like JGEL~\cite{Zhang_Tan_Lu_Yan_Wang_2026}. Furthermore, our method achieves the highest F1 score of 90.46\% on PA100K. These metrics jointly demonstrate that UAPAR maintains a highly balanced and comprehensive recognition capability across diverse pedestrian attributes, consistently outperforming recent strong discriminative baselines.

\paragraph{Effectiveness in Attribute Discovery} In terms of instance-level evaluation, UAPAR exhibits exceptionally high Recall, reaching 93.09\% on PA100K and 91.16\% on PETA, ranking second only to JGEL. This strong positive attribute mining capability is rigorously validated on the heavily imbalanced RAP datasets, yielding highly competitive Recall scores of 87.05\% on RAPv1 and 86.51\% on RAPv2, both ranking second only to JGEL. While generative and hierarchical models like SequencePAR~\cite{jin2025sequenceparunderstandingpedestrianattributes} maintain an advantage in exact-match metrics such as overall Accuracy, Precision, and F1 scores on the RAP benchmarks, UAPAR effectively trades off certain precision to maximize positive attribute localization. The prominent Recall scores indicate that our method is particularly effective at extracting and discovering positive attributes within complex scenes, ensuring lower missed detections while maintaining strong overall performance.

\paragraph{Robustness in Zero-Shot Scenarios} As shown in Table \ref{tab:3}, under zero-shot settings, UAPAR demonstrates superior generalization by achieving the highest mean Accuracy (80.77\% and 81.09\%) and F1 scores (76.78\% and 80.32\%) on the PETA-ZS and RAP-ZS datasets. Although JGEL maintains higher instance-level Recall, our approach provides a more effective precision-recall trade-off, ensuring better overall predictive balance across unseen attribute distributions.

\begin{figure}[t]
    \centering
    \includegraphics[width=\linewidth]{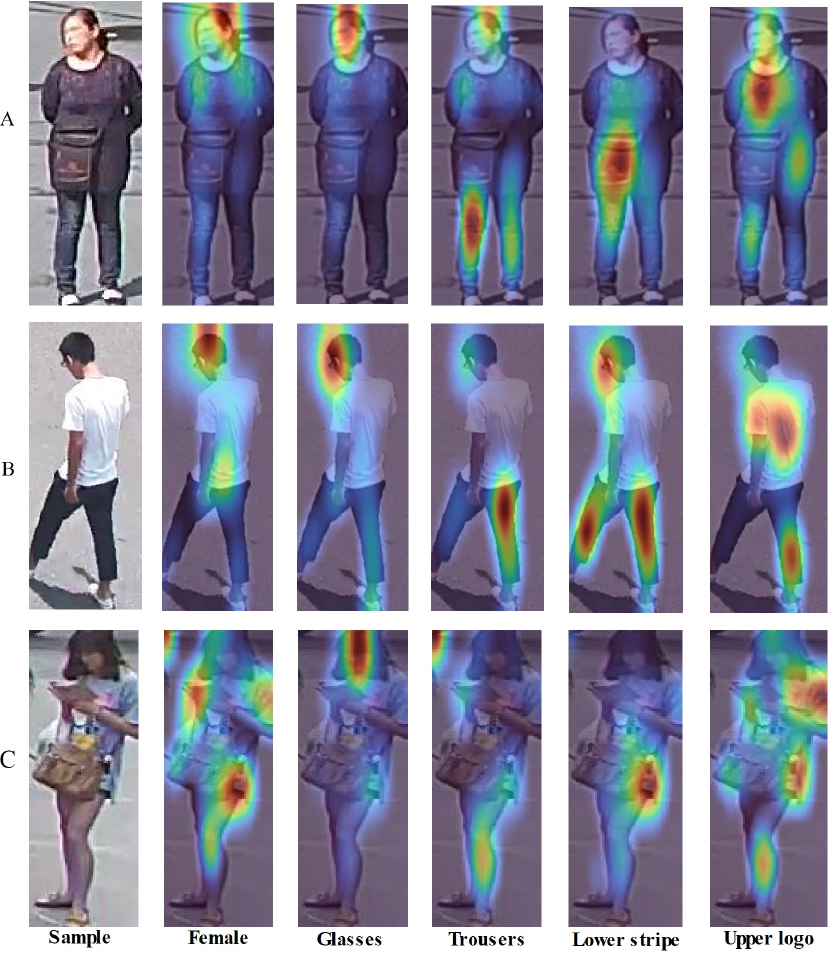}
    \caption{Visualization of the response maps between pedestrian attributes and visual images. The red and blue colors denote the high and low response.}
\label{fig:hot}
\end{figure}

\begin{figure}[t]
    \centering
    \includegraphics[width=\linewidth]{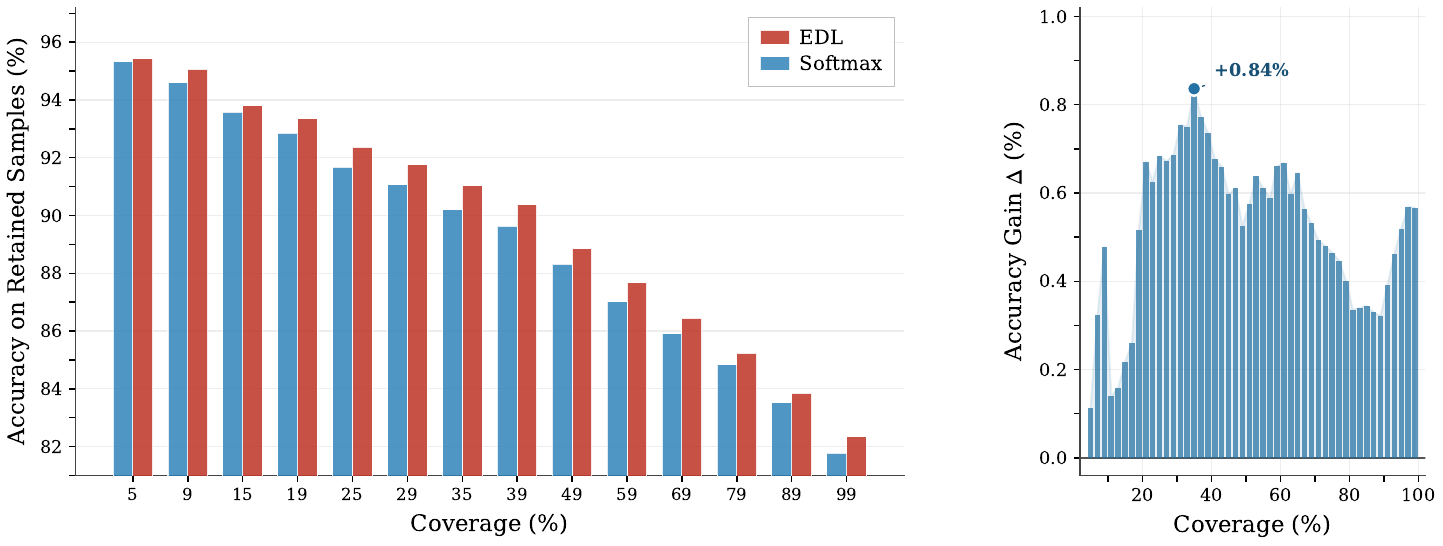}
    \caption{Accuracy-rejection analysis comparing our proposed EDL head against a standard Softmax baseline. EDL consistently achieves higher accuracy on retained samples across all coverage thresholds.}
\label{fig:abro}
\end{figure}
\input{table/ablation}

\subsection{Ablation Studies}
As detailed in Table \ref{tab:4}, starting from the CLIP baseline of 85.50\% mA on PA100K and 87.45\% on PETA, the standalone EDL module provides marginal gains, primarily due to the inherent optimization challenges and convergence difficulties associated with evidential deep learning. However, its true potential is unlocked when coupled with CL, which facilitates model training and enhances feature discriminability, leading to a substantial mA increase to 86.81\% and 88.78\% respectively. By further incorporating RAER and AWR to address localized features and adaptive weighting, the complete framework achieves peak performance with 88.48\% mA on PA100K and 90.74\% mA on PETA, demonstrating that the structural synergy between these modules effectively overcomes training bottlenecks to establish a robust and superior feature representation.

\input{table/SPM}

Furthermore, as shown in Table \ref{tab:5}, the baseline configuration without the Spatial Prior Mask (SPM) achieves 88.54\% mA and 87.60\% F1 on the PETA dataset. Incorporating SPM to inject human structural priors for precise localization guides the model to focus on critical regions, enhancing feature discriminability and improving both metrics to 89.72\% mA and 88.35\% F1.

\begin{figure*}[t]
    \centering
    \includegraphics[width=\linewidth]{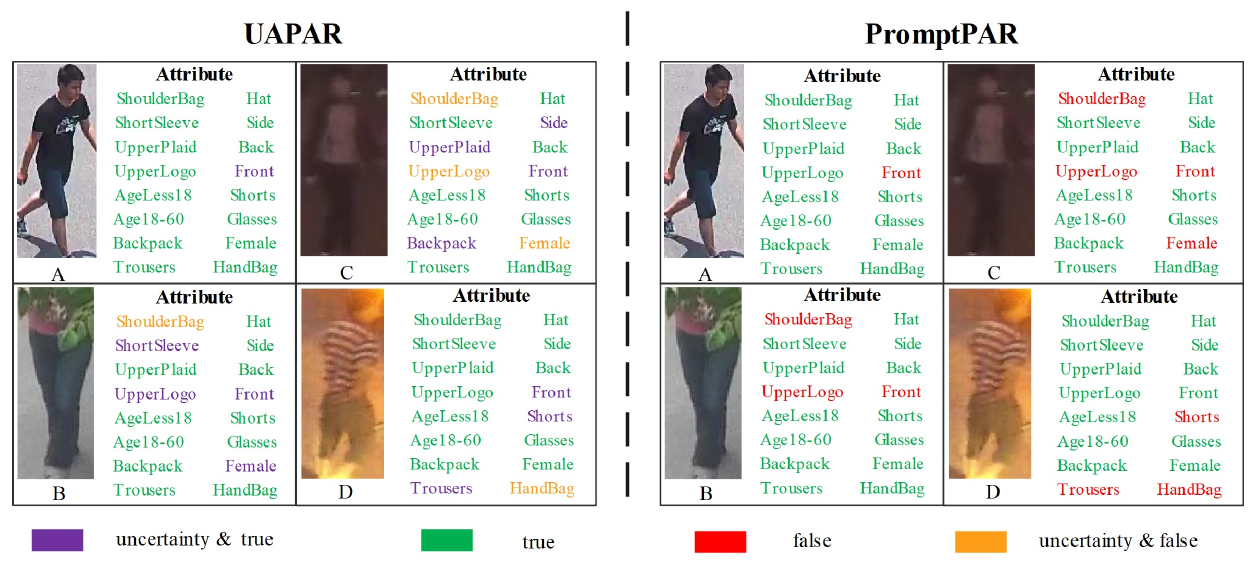}
    \caption{Qualitative comparison of attribute recognition between our UAPAR on the left and the traditional model on the right. \textcolor{green}{Green} and \textcolor{red}{red} denote confident true and false predictions. \textcolor{mypurple}{Purple} and \textcolor{orange}{Orange} indicate true and false results with estimated uncertainty exceeding 0.4.}
    
\label{fig:unc}
\end{figure*}

\subsection{Feature Visualization Analysis}
Figure ~\ref{fig:hot} visualizes the response maps of various attributes, where warmer colors like orange and red denote higher attention and cooler colors indicate lower attention. As shown, highly activated regions accurately align with their corresponding semantic areas. For example, the network correctly focuses on the head for "Glasses" and "Female", and precisely locates the lower body and chest for "Trousers" and "Upper logo", respectively. This demonstrates that our method effectively extracts fine-grained features for Region-Aware Evidence Reasoning, grounding predictions in correct visual evidence rather than background noise.


\subsection{Reliability Analysis of Uncertainty}

\paragraph{Quantitative Accuracy-Rejection Analysis.} To rigorously validate the reliability of our estimated uncertainty, we evaluate the accuracy of retained samples under varying coverage thresholds. As illustrated in Figure~\ref{fig:abro}, our EDL approach consistently outperforms the standard Softmax baseline at all coverage levels, achieving a peak accuracy gain of +0.84\%. This quantitative evidence demonstrates that explicitly quantifying vacuity uncertainty via EDL yields a well-calibrated ranking for filtering challenging samples, effectively mitigating the overconfidence issue of traditional Softmax.

\paragraph{Qualitative Visualizations.} As illustrated in Figure~\ref{fig:unc}, our proposed uncertainty measure effectively identifies attributes under challenging visual conditions, such as severe occlusion or motion blur. Specifically, attributes with high uncertainty scores closely align with visually ambiguous regions. Instead of forcing a hard prediction on these regions which frequently leads to erroneous results, our framework successfully flags them. This visualization demonstrates that the estimated vacuity uncertainty accurately reflects the internal predictive confidence of the model, further proving its effectiveness in filtering out inherently ambiguous samples to maintain overall recognition accuracy.

%% file: table/compare_with_sota.tex
\begin{table*}[t]
\scriptsize
\caption{Comparison with state-of-the-art methods on PA100K and PETA datasets. The best result is \textbf{bolded} and the second best is \underline{underlined}.}
\label{tab:comparison_pa100k_peta}
\centering
\resizebox{\textwidth}{!}{
\begin{tabular}{l|l|ccccc|ccccc}
\hline
\multirow{2}{*}{Methods} & \multirow{2}{*}{Publish}
  & \multicolumn{5}{c|}{PA100K} & \multicolumn{5}{c}{PETA} \\
\cline{3-7} \cline{8-12}
  & & mA & Acc & Prec & Rec & F1 & mA & Acc & Prec & Rec & F1 \\
\hline
DAFL~\cite{article}              & AAAI22  & 83.54 & 80.13 & 87.01 & 89.19 & 88.09 & 87.07 & 78.88 & 85.78 & 87.03 & 86.40 \\
VTB~\cite{9782406}               & TCSVT22 & 83.72 & 80.89 & 87.88 & 89.30 & 88.21 & 85.31 & 79.60 & 86.76 & 87.17 & 86.71 \\
IAA~\cite{WU2022108865}          & PR22    & 81.94 & 80.31 & 88.36 & 88.01 & 87.80 & 85.27 & 78.04 & 86.08 & 85.80 & 85.64 \\

AttExpIB-Net~\cite{10238746}   & TIFS23 & 83.23 & 79.42 & -     & 88.60 & 87.23 & 85.90 & 77.58 & -     & 86.36 & 85.32 \\
EALC ~\cite{WENG2023140} & Neurocomputing23 & 80.52 & 80.13 & -     & 88.59 & 87.88 & 85.94 & 80.58 & -     & 87.38 & 87.44 \\
ViT-RE~\cite{10812850}      & TMM24   & -     & 81.47 & \underline{89.78} & 89.77 & 88.88 & -     & 81.64 & 88.59 & 88.82 & 88.70 \\
PromptPAR~\cite{wang2024pedestrianattributerecognitionclip}       & TCSVT24 & 87.47 & 83.78 & 89.27 & 91.70 & 90.15 & 88.76 & 82.84 & 89.04 & 89.45 & 89.18 \\
SOFA~\cite{Wu_Huang_Gao_Niu_Yang_Gao_Zhao_2024}                 & AAAI24  & 83.40 & 81.10 & 88.40 & 89.00 & 88.30 & 87.10 & 81.10 & 87.80 & 88.40 & 87.80 \\
SSPNet~\cite{SHEN2024110194}  & PR24 & 83.58 & 80.63 & -     & 89.32 & 88.55 & 88.37 & 82.00 & -     & 90.55 & \underline{89.50} \\
HDFL~\cite{WU2025107463}         & NN25    & 84.92 & 80.23 & 87.45 & 88.74 & 87.72 & 87.55 & 79.66 & 87.08 & 87.16 & 86.85 \\
SequencePAR~\cite{jin2025sequenceparunderstandingpedestrianattributes} & PR25 & - & \underline{83.94} & \textbf{90.38} & 90.23 & 90.10 & - & \textbf{84.92} & \textbf{90.44} & 90.73 & \textbf{90.46} \\
FOCUS~\cite{an2025focusfinegrainedoptimizationsemantic}           & ICME25  & 83.90 & 81.23 & 89.29 & 89.87 & 88.41 & 88.04 & 81.96 & 88.56 & 89.07 & 88.54 \\
KGPAR~\cite{wang2025pedestrianattributerecognitionhierarchical}   & arXiv25 & \underline{87.95} & \textbf{84.01} & 89.55 & 91.65 & \underline{90.28} & 88.50 & \underline{83.27} & \underline{89.74} & 89.56 & 89.43 \\
JGEL~\cite{Zhang_Tan_Lu_Yan_Wang_2026}                           & AAAI26  & 87.61 & 82.03 & -     & \textbf{94.91} & 88.92 & \underline{90.27} & 83.23 & -     & \textbf{93.64} & 89.22 \\
\hline

\rowcolor{baselinerow}
UAPAR & Ours & \textbf{88.48} & 82.83 & 87.97 & \underline{93.09} & \textbf{90.46} & \textbf{90.74} & 81.46 & 87.05 & \underline{91.16} & 89.06 \\
\hline
\end{tabular}
}
\label{tab:1}
\end{table*}

\begin{table*}[t]
\scriptsize
\caption{Comparison with state-of-the-art methods on RAPv1 and RAPv2 datasets. The best result is \textbf{bolded} and the second best is \underline{underlined}.}
\label{tab:comparison_rap}
\centering
\resizebox{\textwidth}{!}{
\begin{tabular}{l|l|ccccc|ccccc}
\hline
\multirow{2}{*}{Methods} & \multirow{2}{*}{Publish}
  & \multicolumn{5}{c|}{RAPv1} & \multicolumn{5}{c}{RAPv2} \\
\cline{3-7} \cline{8-12}
  & & mA & Acc & Prec & Rec & F1 & mA & Acc & Prec & Rec & F1 \\
\hline
DAFL~\cite{article}              & AAAI22  & 83.72 & 68.18 & 77.41 & 83.39 & 80.29 & 81.04 & 66.70 & 76.52 & 82.39 & 79.13 \\
VTB~\cite{9782406}               & TCSVT22 & 82.67 & 69.44 & 78.28 & 84.39 & 80.84 & 81.34 & 67.48 & 76.41 & 83.32 & 79.35 \\
IAA~\cite{WU2022108865}          & PR22    & 81.72 & 68.47 & 79.56 & 82.06 & 80.37 & 79.99 & 68.03 & \underline{78.75} & 81.37 & 79.69 \\

AttExpIB-Net~\cite{10238746}   & TIFS23 & 82.46 & 68.81 & -     & 81.63 & 80.25 & 80.60 & 67.31 & -     & 80.38 & 79.15 \\
EALC ~\cite{WENG2023140} & Neurocomputing23 & 82.69 & 69.30 & -     & 82.77 & 81.17 & -     & -     & -     & -     & -     \\
ViT-RE~\cite{10812850}      & TMM24   & -     & 69.45 & \underline{81.18} & 80.80 & 80.99 & -     & -     & -     & -     & -     \\
PromptPAR~\cite{wang2024pedestrianattributerecognitionclip}       & TCSVT24 & 85.45 & 71.61 & 79.64 & 86.05 & 82.38 & 83.14 & 69.62 & 77.42 & 85.73 & 81.00 \\
SOFA~\cite{Wu_Huang_Gao_Niu_Yang_Gao_Zhao_2024}                 & AAAI24  & 83.40 & 70.00 & 80.00 & 83.00 & 81.50 & 81.90 & 68.60 & 78.00 & 83.10 & 80.20 \\
SSPNet~\cite{SHEN2024110194}  & PR24 & 83.24 & 70.41 & -     & 82.49 & 81.90 & -     & -     & -     & -     & -     \\
HDFL~\cite{WU2025107463}         & NN25    & 83.68 & 70.49 & 80.25 & 83.55 & 81.51 & 81.81 & 68.22 & 78.30 & 81.28 & 79.85 \\
SequencePAR~\cite{jin2025sequenceparunderstandingpedestrianattributes} & PR25 & - & 71.47 & \textbf{82.40} & 82.09 & 82.05 & - & \textbf{70.14} & \textbf{81.37} & 81.22 & 81.10 \\
FOCUS~\cite{an2025focusfinegrainedoptimizationsemantic}           & ICME25  & 83.45 & 70.14 & 80.10 & 85.18 & \underline{82.41} & -     & -     & -     & -     & -     \\
KGPAR~\cite{wang2025pedestrianattributerecognitionhierarchical}   & arXiv25 & 85.05 & \underline{71.75} & 79.95 & 85.98 & \textbf{82.50} & 83.02 & \underline{70.03} & 78.33 & 85.17 & \underline{81.27} \\
JGEL~\cite{Zhang_Tan_Lu_Yan_Wang_2026}                           & AAAI26  & \underline{86.09} & \textbf{72.23} & -     & \textbf{90.64} & 81.32 & \underline{85.54} & 68.24 & -     & \textbf{91.73} & 79.79 \\
\hline

\rowcolor{baselinerow}
UAPAR & Ours & \textbf{87.48} & 70.81 & 78.09 & \underline{87.05} & 81.91 & \textbf{85.71} & 67.93 & 74.87 &  \underline{86.51} & \textbf{82.33} \\
\hline
\end{tabular}
}

\label{tab:2}
\end{table*}

%% file: table/zero_shot.tex
\begin{table}[htbp]
  \centering
  \caption{Comparison with several state-of-the-art methods on the PETA-ZS and RAP-ZS datasets.}
  \resizebox{\linewidth}{!}{
    \setlength{\tabcolsep}{1.2mm}
    \begin{tabular}{c|l|cccc}
    \hline \hline
    Data & Methods & mA & Acc & Recall & F1 \\
    \hline
    \multirow{6}{*}{\rotatebox{90}{PETA-ZS}} 
    & SOFAFormer ~\cite{Wu_Huang_Gao_Niu_Yang_Gao_Zhao_2024} & 74.70 & 62.10 & 75.10 & 74.60 \\
    & AAR ~\cite{WU2025130236} & - & 62.89 & 75.66 & 75.38 \\
    & HDFL ~\cite{WU2025107463} & - & 62.01 & 75.36 & 74.78 \\
    
    & JGEL ~\cite{Zhang_Tan_Lu_Yan_Wang_2026} & \underline{80.56} & \textbf{64.98} & \textbf{86.39} & \underline{76.75} \\
    \cline{2-6}
    & \textbf{UAPAR (Ours) } & \textbf{80.77} & \underline{64.77} & \underline{79.63} & \textbf{76.78} \\
    \hline
    \multirow{6}{*}{\rotatebox{90}{RAP-ZS}} 
    & SOFAFormer ~\cite{Wu_Huang_Gao_Niu_Yang_Gao_Zhao_2024} & 73.90 & 66.30 & 79.40 & 78.40 \\
    & AAR ~\cite{WU2025130236} & - & 66.51 & 79.13 & 78.60 \\
    & HDFL ~\cite{WU2025107463} & - & 66.70 & 79.81 & 78.42 \\
    
    & JGEL ~\cite{Zhang_Tan_Lu_Yan_Wang_2026} & \underline{80.10} & \underline{67.96} & \textbf{90.15} & \underline{79.73} \\
    \cline{2-6}
    & \textbf{UAPAR (Ours) }& \textbf{81.09} & \textbf{68.60} & \underline{86.33} & \textbf{80.32} \\
    \hline \hline
    \end{tabular}
  }
\label{tab:3}
\end{table}

%% file: table/ablation.tex
\definecolor{baselinerow}{gray}{0.9} 

\begin{table}[t]
\centering
\caption{Component ablation on PA100K and PETA. All configurations use CLIP ViT-L/14 with prompt tuning as the backbone.}
\label{tab:ablation}
\setlength{\tabcolsep}{3.5pt}
\resizebox{\columnwidth}{!}{%
\begin{tabular}{c|c|cccc|cc|cc}
\hline\hline
\multirow{2}{*}{No.} & \multirow{2}{*}{CLIP} & \multirow{2}{*}{EDL} & \multirow{2}{*}{CL} & \multirow{2}{*}{RAER} & \multirow{2}{*}{AWR} & \multicolumn{2}{c|}{PA100K} & \multicolumn{2}{c}{PETA} \\ \cline{7-10}
 & & & & & & mA & F1 & mA & F1 \\
\hline
1 & \checkmark & $\times$ & $\times$ & $\times$ & $\times$ & 85.50 & 88.70 & 87.45 & 87.93 \\
2 & \checkmark & \checkmark & $\times$ & $\times$ & $\times$ & 85.78 & 88.81 & 87.52 & 87.40 \\
3 & \checkmark & \checkmark & \checkmark & $\times$ & $\times$ & 86.81 & 89.23 & 88.78 & 88.61 \\
4 & \checkmark & \checkmark & $\times$ & \checkmark & $\times$ & 87.09 & 89.47 & 89.59 & 88.15 \\
5 & \checkmark & \checkmark & \checkmark & \checkmark & $\times$ & 87.27 & 89.38 & 89.72 & 88.35 \\
6 & \checkmark & \checkmark & $\times$ & $\times$ & \checkmark & 87.78 & 89.21 & 89.46 & 88.01 \\
\rowcolor{baselinerow}
7 & \checkmark & \checkmark & \checkmark & \checkmark & \checkmark & \textbf{88.48} & \textbf{90.46} & \textbf{90.74} & \textbf{89.06} \\
\hline\hline
\end{tabular}%
}
\label{tab:4}
\end{table}

%% file: table/SPM.tex
\begin{table}[t]
\centering
\small
\caption{Effect of SPM on the base configuration.}
\label{tab:5}
\setlength{\tabcolsep}{10pt}
\renewcommand{\arraystretch}{1.15}
\begin{tabular}{l cc}
\toprule
\textbf{RAER} & \textbf{mA} & \textbf{F1} \\
\midrule
w/o\hspace{1pt} SPM & 88.54 & 87.60 \\
w/\phantom{o}\hspace{1pt} SPM & \textbf{89.72} & \textbf{88.35} \\
\bottomrule
\end{tabular}
\vspace{-6pt}
\end{table}

%% file: latex/conclusion.tex
In this paper, we present UAPAR, an EDL-based uncertainty-aware framework for PAR. By combining the RAER module with an uncertainty-guided curriculum learning strategy, our method achieves state-of-the-art performance on multiple benchmarks and produces reliable uncertainty estimates for challenging samples. Although improvements are still needed for extreme few-shot attributes in long-tail distributions, our results highlight the potential of evidential reasoning in enhancing the robustness and interpretability of PAR systems. Future work will explore integrating advanced imbalanced learning methods with evidential modeling to better recognize rare attributes.

%% file: latex/acknowledge.tex
This work was supported by the National Natural Science Foundation of China (No. 62301346), the Science and Technology Program of the Tibet Autonomous Region, China (No. XZ202502YD0003), the Open Project Program of The Key Laboratory of Cognitive Computing and Intelligent Information Processing of Fujian Education Institutions, Wuyi University (No. KLCCIIP202403)